# Perfectly Accurate Membership Inference by a Dishonest Central Server in Federated Learning

Georg Pichler, *Member, IEEE,* Marco Romanelli, Leonardo Rey Vega, *Member, IEEE,* and Pablo Piantanida, *Senior Member, IEEE*

**Abstract**—Federated Learning is expected to provide strong privacy guarantees, as only gradients or model parameters but no plain text training data is ever exchanged either between the clients or between the clients and the central server. In this paper, we challenge this claim by introducing a simple but still very effective membership inference attack algorithm, which relies only on a single training step. In contrast to the popular honest-but-curious model, we investigate a framework with a dishonest central server. Our strategy is applicable to models with ReLU activations and uses the properties of this activation function to achieve perfect accuracy. Empirical evaluation on visual classification tasks with MNIST, CIFAR10, CIFAR100 and CelebA datasets show that our method provides perfect accuracy in identifying one sample in a training set with thousands of samples. Occasional failures of our method lead us to discover duplicate images in the CIFAR100 and CelebA datasets.

---

## 1 Introduction

When operating on user data, a fundamental advantage of Federated Learning (FL), compared to centralized learning, is the privacy gain experienced by the user. No plain text training data, but only gradients or model parameters are exchanged with a Central Server (CS). However, a dishonest CS can still mount attacks against its users. In this work, we study the privacy risk for a participant in a FL system, that results from a dishonest CS. At the same time, this scenario is also applicable to an active man-in-the-middle attacker between the client and the CS.

To motivate the study of this particular membership inference attack problem, we provide a fictitious, but plausible scenario, where such an attack could be successfully mounted:

The popular mobile phone manufacturer A wants to improve the operating systems of their phones by training a superior image recognition software, allowing their customers to easily categorize snapshots. In order to do this, company A employs FL and obtains consent of (most of) its users for training on the customer's device using their local photo library. Thus, A sends a parameter vector $\theta$ of a neural network to each phone. Without user interaction, the mobile phone trains the neural network on the local photo library and communicates the result to A. Crucially, the images never leave the device, providing customers with a sense of privacy.

At the time, in a kidnapping case, a photo of the victim is sent to the police. Suspecting that the photo may have been taken with a device produced by A, the police approach A, asking if it is possible to determine which user has this particular image on their mobile phone. In this paper, we show how A can craft a special set of parameters $\theta$ in order to accomplish the police request. Letting each user train on $\theta$ for one iteration, A can deduce which user has this particular image in their photo library, by evaluating the answers. The impact on the FL system is negligible as only one iteration of training was needed, but A can achieve perfectly accurate membership inference as we will show in this work.

### 1.1 Privacy in FL

In a FL setup, a CS trains a model by communicating with one or more clients who hold training data. Letting $\theta$ denotes model parameters, typically FL proceeds as depicted in Algorithm 1: With selectSubset a (possibly random) subset of clients is se-

---

**Algorithm 1:** FL overview.

> **input** : initial parameters $\theta_0$; number of iterations $I$; set of clients $\mathscr{C}$.
> **output:** final parameters $\theta_I$.
> answers ← []
> for $i \in \{1, \ldots, I\}$ do
>     $\mathscr{A}$ ← selectSubset($\mathscr{C}$)
>     for client $\in \mathscr{A}$ do
>         answers.append(client($\theta_{i-1}$))
>     end
>     $\theta_i$ ← aggregate(answers)
> end
> return $\theta_I$

---

lected for the current iteration. Every client in this subset is then queried and the results stored. Before the next iteration begins, an aggregation function `aggregate` is used to obtain the next parameter vector. The data returned by the clients, as well as the aggregation function vary, depending on the FL algorithm. E.g., in Federated Stochastic Gradient Descent (FedSGD) [1], clients compute the (average) gradient on a local mini-batch and the CS takes the arithmetic average of these gradients and performs a gradient descent step. Similarly, in Federated Averaging (FedAvg) [2], clients train on several batches, possibly for multiple epochs, returning the full parameter vector. The CS uses the arithmetic average of these parameter vectors for the next iteration.

We are interested in the privacy aspect of this setup from the client's point of view. In particular, we want to study the

- *This research was supported by DATAIA "Programme d'Investissement d'Avenir" (ANR-17-CONV-0003).*
- *G. Pichler is with Technische Universität Wien, 1040 Vienna, Austria. Email: georg.pichler@ieee.org.*
- *Marco Romanelli is with New York University, Tandon School of Engineering, 370 Jay Street, Brooklyn, NY 11201, USA. Email: mr6852@nyu.edu.*
- *Leonardo Rey Vega is with CSC-CONICET and Universidad de Buenos Aires, C1425FQD Buenos Aires, Argentina. Email: lrey@fi.uba.ar.*
- *Pablo Piantanida is with ILLS · International Laboratory on Learning Systems and MILA - Quebec AI Institute, CNRS, CentraleSupélec, Montreal, QC H3C 1K3, Canada. Email: piantani@mila.quebec.*

membership inference problem, where the CS wants to learn whether a particular training sample is present in the training data of a particular client. From Algorithm 1 it is clear that the only interaction between a particular client and the CS is through the function $\phi = \text{client}(\theta)$, where the CS provides a parameter vector $\theta$ and retrieves the answer $\phi$. Thus, we will solely focus on this interaction between the CS and one particular client.

Moreover, we perform our attack using only one query, i.e., the CS determines membership based only on the output $\phi = \text{client}(\theta)$ for one input $\theta$.

### 1.2 Previous Work

Privacy of FL has been an active area of research since this field developed. We focus on privacy at training time and in particular the task of membership inference conducted by the CS. Most literature either assumes a client-side attacker [3], [4] or, that a malicious CS behaves honest-but-curious [5], [6], [7]. Much work has also been devoted to reconstructing input from gradients [4], [8], [9]. This problem is rather trivial if the gradient at a single input is available, but becomes challenging when a whole batch is processed, the typical scenario in FedSGD, where each client reports its gradient computed on one mini-batch to the CS.

An active attacker is studied in [10], [11], as well as the recent works [12], [13], [14], [15]. The focus of these works is the reconstruction of individual data samples in the client dataset when the FedSGD strategy is used. There, the full gradients are shared with the CS. Only [13, Sec. 4.3] makes an attempt at the FedAvg setting, but uses a very limited batchsize, number of batches and epochs. The method employed in [15, Sec. VI] is similar to the strategy employed in our work, relying on the conditional activation and deactivation of parts of the network.

The paper [16] also studies membership inference in federated learning, displaying remarkable accuracy. It, however, requires several rounds of communication, while our strategy achieves perfect accuracy in the active-adversary scenario with only one round.

In [11], an active, global adversary is considered. However, the attacker at the CS in [11] requires additional training for several iterations and a significant amount of additional training data. This much more, than what is afforded to the attacker in this work, where the CS may only query the client once and, apart from the one sample to be tested, does not have any additional data.

Additionally, there is vast literature on the privacy of machine learning models in general [17], [18], [19]. However, these works assume a fully trained model and tend to exploit overfitting to be able to perform membership inference. The scenario considered here is different in that the attack is only performed during one iteration.

A different line of research considers defense mechanisms such as encrypted parameter updates and homomorphic encryption schemes [20], [21], [22].

## 2 Our Method

### 2.1 Setup and Attacker Model

Our method relies crucially on the following assumptions.

1) The model $f(x; \theta)$ uses ReLU activations and the structure depicted in Fig. 1 can be embedded in f.
2) The client uses a form of stochastic gradient descent, e.g., SGD, Adam [23] to minimize a loss function.
3) The function client(), i.e., all training hyperparameters (batch size, optimizer, ...) are known to the CS.

While certainly adaptable to other settings, we will detail and showcase our strategy for FedAvg in the task of image classification. Thus, we will additionally assume the following.

4) The dataset $\mathscr{D} = \{(x_n, y_n)\}_{n=1,...,N}$ consists of images $x_n$ and discrete labels $y_n \in \{1, ..., L\}$.
5) The model outputs logits and we use cross-entropy with softmax, i.e., the client attempts to minimize the loss function $\ell$ at the training samples $\mathscr{D}$, where

$$\ell(x, y; \theta) = -\log \frac{\exp(f(x; \theta)_y)}{\sum_{l=1}^{L} \exp(f(x; \theta)_l)} \qquad (1)$$

$$= -f(x; \theta)_y + \log \sum_{l=1}^{L} \exp(f(x; \theta)_l). \qquad (2)$$

6) FedAvg is used, i.e., the response $\phi$ to the CS is the full parameter vector after training. The client trains for $E$ epochs using $J$ mini-batches with batch size $B = \frac{N}{J}$.
7) The structure depicted in Fig. 1 can be embedded in the last layers of f, obtaining $b$ as a logit in the output.

The CS is given the target sample $s = (x_t, y_t)$, where $t \in \{0, 1\}$. Note that we have $s \in \mathscr{D}$ if $t = 1$, while for $t = 0$, $s = (x_0, y_0)$ is another sample from the dataset, not present in $\mathscr{D}$. Using $s$, the CS needs to produce a parameter vector $\theta$, which is used by the client to compute the response $\phi = \text{client}(\theta, \mathscr{D})$. This response is used by the CS, to obtain an estimate $\hat{T}$ of $t$. If $\hat{T} = t$ the CS correctly identified membership.

**Remark 2.1.** *Note, that FedAvg is more challenging for an attacker, as FedSGD is essentially a special case of FedAvg. In order to see this, assume that FedAvg is used, with $E = 1$ epoch and $J = 1$ mini-batch, i.e., $B = N$. Then the entire training set $\mathscr{D}$ consists of only one single mini-batch. The difference $\theta - \phi$ between the parameter vectors is simply the (scaled) gradient of (1) at $\mathscr{D}$.*

**Remark 2.2.** *Note that the dishonest server model used here is equivalent to a man-in-the-middle between the CS and the client with the capability to intercept and manipulate traffic.*

*The CS has no access to client data nor are additional training samples (apart from the target sample s) required. However, we do require the optimizer and the batch size to be known to the CS. But as the CS needs access to the model architecture by design, assuming that other hyper-parameters are also known to the CS seems reasonable. After all, it is very likely that the same entity/company that operates the CS also provides the client application. Otherwise, these parameters may also be determined by reverse-engineering of the client application or simply by trial-and-error.*

**Remark 2.3.** *Our method crucially relies on Assumption 1, i.e., it is possible to choose parameters $\theta$ of $f(x, \theta)$, such that the network depicted in Fig. 1 can be realized inside f. For the sake of presentation we assume that this embedding is possible in the final layers (Assumption 7), which simplifies the calculation of derivatives, but is not strictly necessary for our method to be applicable.*

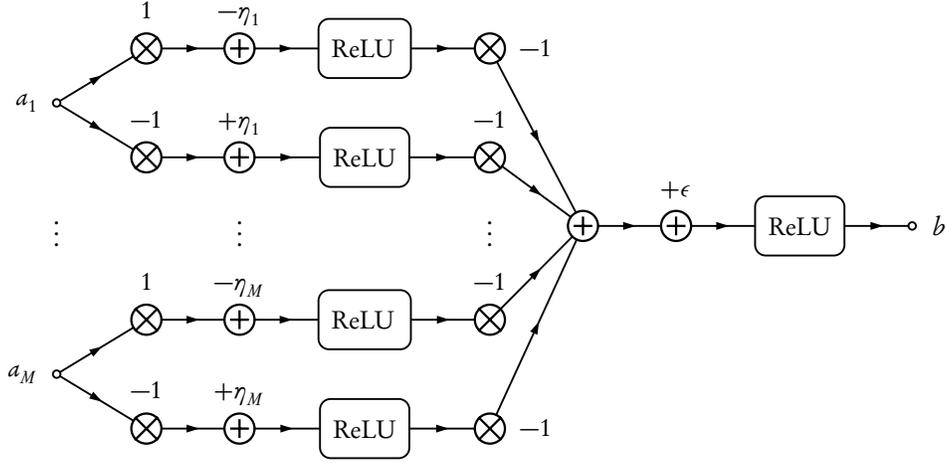

Fig. 1. Part of a neural network for the identification of $M$ values $a_1, a_2, \ldots, a_M$. Multiplication with weights and addition of biases are depicted as separate operations. The weights for the neurons in the first hidden layer are $\pm 1$, while their biases are $\mp \eta_m$. The weights for the neuron in the second hidden layer are $-1$ and its bias is $\epsilon$.

## 2.2 The Strategy

Before we detail the construction of the model parameters $\theta$, that the CS uses to determine membership of the target sample $s$ in $\mathscr{D}$, we want to provide the main idea behind it. Essentially, we embed the structure shown in Fig. 1 in the clients model. Each hidden node of this structure has a ReLU activation function. It is designed, such that the derivative of its output w.r.t. all parameters in this structure is zero, unless all the inputs $a_j$ are close to $\eta_j$. By choosing $\eta_j$ equal to the values $a_j$ produced by $s$, we can use this property to test for membership: If $s \notin \mathscr{D}$ and no other sample randomly falls within this narrow interval, the initial parameters will not change during training.

The following lemma summarizes the central properties of the network in Fig. 1.

**Lemma 2.1.** *Given the network depicted in Fig. 1, we have*[1]

$$b = \mathrm{ReLU}\left(\epsilon - \sum_{m=1}^{M} |a_m - \eta_m|\right) \text{ and} \tag{3}$$

$$\partial_\epsilon b = \mathbb{1}_{[0,\epsilon]}\left(\sum_{m=1}^{M} |a_m - \eta_m|\right). \tag{4}$$

*Furthermore, for any parameter $\theta'$ (weight or bias) in the hidden layers of the network depicted in Fig. 1, we have $\partial_{\theta'} b = 0$ if $\sum_{m=1}^{M} |a_m - \eta_m| > \epsilon$.*

*Proof.* We have

$$b = \mathrm{ReLU}\left(\epsilon - \sum_{m=1}^{M} \mathrm{ReLU}(a_m - \eta_m) + \mathrm{ReLU}(-a_m + \eta_m)\right)$$
$$= \mathrm{ReLU}\left(\epsilon - \sum_{m=1}^{M} |a_m - \eta_m|\right), \tag{5}$$

as in general $\mathrm{ReLU}(z) + \mathrm{ReLU}(-z) = |z|$. For $\sum_{m=1}^{M} |a_m - \eta_m| \leq \epsilon$, we have $b(\epsilon) = \epsilon - \sum_{m=1}^{M} |a_m - \eta_m|$ and thus $\partial_\epsilon b(\epsilon) = 1$. Let $\theta'$ denote any parameter, weight or bias, in the hidden layers of the network in Fig. 1. If $\sum_{m=1}^{M} |a_m - \eta_m| > \epsilon$, we have $c = \epsilon - \sum_{m=1}^{M} |a_m - \eta_m| < 0$. As $c(\theta')$ is continuous, there is a neighborhood of $\theta'$ with $c < 0$. Thus, $b = \mathrm{ReLU}(c)$ is locally constant (and equal to zero) at $\theta'$, yielding $\partial_{\theta'} b = 0$. □

Given a target sample $s = (x_t, y_t)$, the CS obtains $\theta$ as follows.

1) The model f consists of two parts $\mathrm{f} = \mathrm{f}_1 \circ \mathrm{f}_0$ where the structure in Fig. 1 is embedded in $\mathrm{f}_1$ (cf. Assumption 7). Initialize the parameters $\theta_0$ of $\mathrm{f}_0$ randomly. In the parameter vector $\theta_1$ of $\mathrm{f}_1$, initialize all weights with zeros and all biases with $-1$.

2) Let the $M$ largest (in absolute value) components of $\mathrm{f}_0(x_t)$ be $(a_1, \ldots a_M)$ and their value be $(\eta_1, \ldots, \eta_M)$.

3) Use these $M$ components $(a_1, \ldots a_M)$ as inputs to the model in Fig. 1 and set the corresponding weights and biases in $\theta_1$. The output $b$ corresponds to the $y_t$-th component of the output.

As all but the $y_t$-th output are zero, we use (1) to obtain the loss function

$$\ell(x, y; \theta) = -b \, \mathbb{1}_{\{y_t\}}(y) + \log(L - 1 + e^b) \tag{6}$$

and by Lemma 2.1 its derivative w.r.t. $\epsilon$ is

$$\partial_\epsilon \ell(x, y; \theta) = [\partial_b \ell(x, y; \theta)] \cdot [\partial_\epsilon b] \tag{7}$$

$$= \mathbb{1}_{[0,\epsilon]}\left(\sum_{m=1}^{M} |a_m - \eta_m|\right)\left(\frac{e^b}{L - 1 + e^b} - \mathbb{1}_{\{y_t\}}(y)\right). \tag{8}$$

Crucially, by Lemma 2.1, the derivative of $\ell$ w.r.t. *all* network parameters in the hidden layers of the network in Fig. 1 is zero if $\sum_{m=1}^{M} |a_m - \eta_m| > \epsilon$. Thus, if all inputs $x_n$ in the training set result in $a_1, \ldots, a_M$ that are not close to $\eta_1, \ldots, \eta_M$, then these parameters remain unchanged during the entire training process. For the decision function, we measure the change the parameter $\epsilon$.

## 2.3 Decision

After obtaining the parameter vector $\phi = \mathrm{client}(\theta, \mathscr{D})$ resulting from the training at the client, the CS needs to make a decision $\hat{T}$. This is performed by a threshold test on the $\epsilon$-component of $\phi$, denoted $\hat{\epsilon}$.

The CS computes one iteration of the optimization procedure with the crafted parameters $\theta$ on the single sample $s = (x_t, y_t)$, obtaining $\dot{\phi} = \mathrm{client}(\theta, \{s\})$. This is possible by Assumption 3. The decision statistic is

$$\Delta = B \frac{|\hat{\epsilon} - \epsilon|}{|\dot{\epsilon} - \epsilon|}, \tag{9}$$

---

1. We use the indicator $\mathbb{1}_{\mathscr{A}}(a) = 1$ if $a \in \mathscr{A}$ and $\mathbb{1}_{\mathscr{A}}(a) = 0$ if $a \notin \mathscr{A}$. The notation $\partial_\epsilon b$ is used to denote the derivative of $b$ w.r.t. $\epsilon$.

where $\hat{\epsilon}$ denotes the $\epsilon$-component of $\hat{\phi}$ and $B$ is the batch size. For a fixed threshold $\xi$, the CS decides for $\hat{T} = 1$ if $\Delta \geq \xi$ and $\hat{T} = 0$ otherwise.

## 3 Experiments

We train a simple network from the tutorial of the FL framework flower [24] with MNIST, CIFAR10, CIFAR100, and CelebA datasets[2].

The simple network architecture is[3]

$$\underbrace{\left(\text{Conv2D} \to \text{ReLU} \to \text{Pool2D}\right)}_{f_0} \to \underbrace{\left(\text{Lin} \to \text{ReLU}\right)^2 \to \text{Lin}}_{f_1}$$

and thus satisfies Assumption 1.

Python code to reproduce our experiments can be found at https://github.com/g-pichler/dishonest_mia.

### 3.1 Training

All parameters are listed in Table 1. For each run, we randomly select $N$ samples $\mathscr{D}$ from the training set (MNIST, CIFAR10, CIFAR100, or CelebA). For half the runs, we randomly pick an element $s \in \mathscr{D}$, for the other half, we randomly pick an element from the remaining samples, thus ensuring $s \notin \mathscr{D}$. We then spawn a `flower`-client and a `flower`-server, implementing our dishonest strategy. Finally, after every run, we record $\Delta$, as given by (9), as well as whether $s \in \mathscr{D}$ or not. A failure of our method occurs if $\Delta \geq \xi$ when $s \notin \mathscr{D}$ (False Positive) or if $\Delta < \xi$ and $s \in \mathscr{D}$ (False Negative).

TABLE 1
Experiment Details.

| Parameter | Value |
|---|---|
| Batch size | $B = 32$ |
| Epochs | $E$ |
| Threshold | $\xi = 0.1$ |
| Number of samples | $N = B \cdot J = 32 \cdot J$ |
| Number of values | $M = 4$ |
| Iterations (batches per epoch) | $J$ |
| Optimizer | SGD, Adam |
| $\epsilon$ | $\epsilon = 10^{-3}$ |
| Runs | 400 |

We perform 400 runs on the training set, using the fixed threshold $\xi = 0.1$.

### 3.2 Results

We repeated our experiment 6 times with $E = 1$ epoch, varying the number of batches $J \in \{1, 4, 16, 64, 128, 256\}$ in the epoch. As each batch consists of $B = 32$ samples, this corresponds to between $N = 32$ and $N = 8192$ samples. In addition, we conducted further experiments with $J = 128$ batches per epoch (i.e., $N = 4096$), varying the number of epochs $E \in \{1, 2, 4\}$.

All experiments concluded without a single false decision by the CS, achieving an over-all accuracy of 100%.

We experienced very occasional failures of our method on the CIFAR100 and CelebA datasets during exploration. When investigating, we noticed that there are 14 exact duplicate images in the training set of the CIFAR100 dataset and 82 exact duplicates in the training set of the CelebA dataset. More details on these duplicates are provided in the discussion (Section 4.1). In addition, there are several near duplicate images present in the CelebA dataset. These images are indistinguishable for a human and large portions of the images are identical. These (near) duplicates lead to very occasional false positives of our method. We did not observe a single failure that could not be attributed to one of these (near) duplicates.

---

2. The datasets can be found at: CIFAR10/100, MNIST, CelebA [25]
3. We use the notation $F^2 = F \to F$ to denote functional composition.

We chose $\epsilon = 0.001$ relatively large to obtain robustness against numerical errors. Then, $M = 4$ is sufficient to achieve perfect accuracy. However, choosing small $M$, e.g., $M \in \{1, 2\}$, will then lead to false positives. This is showcased in the ablation study on $M$ that can be found in Table 2. We provide False Positive Rate (FPR), False Negative Rate (FNR), accuracy (acc.), and the Area under Curve (AUC) of the Receiver Operating Characteristic. All values are rounded to the second place value after the decimal point.

We found our method to be robust to changes of the threshold $\xi$. With $t = 0$, i.e., $x \notin \mathscr{D}$, we found $\Delta = 0.00$ up to machine precision for every sample outside the training set when $M \geq 4$.

For $t = 1$, i.e., $s \in \mathscr{D}$, we observed $\Delta \geq 0.23$ for any sample in the training set and $M = 4$. This prompted us to use the threshold $\xi = 0.1$. However, with $M = 8$ we obtain $\Delta \geq 1.00$ for every sample in the training set. Note that this improvement comes only at a small cost in terms of number of neurons, as the number of neurons scales linearly in $M$. To detect $M$ values, only $2M + 1$ neurons with ReLU activation are required (cf. Fig. 1).

## 4 Discussion and Possible Extensions

We presented a simple algorithm allowing membership inference attacks in a FL setup. Our attack achieves perfect accuracy when identifying a single target sample among thousands of training samples, using only a single query. However, it should still be regarded as a proof-of-concept.

We refrained from performing our experiments on larger models. The model size does not impact our strategy, as most weights would be set to zero. We thus decided not to train larger models in order to keep the computation time to a minimum.

We only performed identification based on up to 16 components as a proof-of-concept. However, if enough neurons are available, an extension to more components is certainly feasible. Note the small size of our embedded network in Fig. 1, compared to the methods of [12], [13], [14], [15]. To identify a sample based on $M$ values, $2M + 1$ hidden neurons suffice. Depending on the architecture, it may be feasible to obfuscate membership inference, while still yielding a model performing well at the target task.

Finally, we want to remark that the strategy presented here may in fact be used for attribute inference. For instance, $M - 1$ known attributes can be used to select a specific target sample and a private attribute can then be inferred using multiple queries.

### 4.1 Duplicates

As our strategy may appear to fail if exact duplicates are present in the dataset, we performed a search for such images in all four datasets (MNIST, CIFAR10, CIFAR100, CelebA). Both MNIST and CIFAR10 are free of exact duplicate images. We (re)discovered the exact duplicates in the CIFAR100 dataset that were pointed out by [26].

TABLE 2
Performance when the client trains for $E = 1$ epoch, $J = 128$ batches per epoch and different $M$.

| Opt. | M | MNIST | | | | CIFAR10 | | | | CIFAR100 | | | | CelebA | | | |
|---|---|---|---|---|---|---|---|---|---|---|---|---|---|---|---|---|---|
| | | FPR | FNR | Acc. | AUC | FPR | FNR | Acc. | AUC | FPR | FNR | Acc. | AUC | FPR | FNR | Acc. | AUC |
| SGD | 1 | 88.00 | 0.00 | 56.00 | 0.66 | 90.00 | 0.50 | 54.75 | 0.58 | 90.50 | 1.00 | 54.25 | 0.91 | 0.00 | 0.50 | 99.75 | 1.00 |
| | 2 | 1.00 | 0.00 | 99.50 | 1.00 | 3.00 | 0.00 | 98.50 | 1.00 | 2.50 | 0.00 | 98.75 | 1.00 | 0.00 | 0.00 | 100.00 | 1.00 |
| | 4 | 0.00 | 0.00 | 100.00 | 1.00 | 0.00 | 0.00 | 100.00 | 1.00 | 0.00 | 0.00 | 100.00 | 1.00 | 0.00 | 0.00 | 100.00 | 1.00 |
| | 8 | 0.00 | 0.00 | 100.00 | 1.00 | 0.00 | 0.00 | 100.00 | 1.00 | 0.00 | 0.00 | 100.00 | 1.00 | 0.00 | 0.00 | 100.00 | 1.00 |
| | 16 | 0.00 | 0.00 | 100.00 | 1.00 | 0.00 | 0.00 | 100.00 | 1.00 | 0.00 | 0.00 | 100.00 | 1.00 | 0.00 | 0.00 | 100.00 | 1.00 |
| Adam | 1 | 83.00 | 0.00 | 58.50 | 0.53 | 88.50 | 0.00 | 55.75 | 0.50 | 84.50 | 0.00 | 57.75 | 0.60 | 47.50 | 0.00 | 76.25 | 0.80 |
| | 2 | 1.50 | 0.00 | 99.25 | 0.99 | 6.00 | 0.00 | 97.00 | 0.96 | 1.50 | 0.00 | 99.25 | 1.00 | 2.50 | 0.00 | 98.75 | 0.99 |
| | 4 | 0.00 | 0.00 | 100.00 | 1.00 | 0.00 | 0.00 | 100.00 | 1.00 | 0.00 | 0.00 | 100.00 | 1.00 | 0.00 | 0.00 | 100.00 | 1.00 |
| | 8 | 0.00 | 0.00 | 100.00 | 1.00 | 0.00 | 0.00 | 100.00 | 1.00 | 0.00 | 0.00 | 100.00 | 1.00 | 0.00 | 0.00 | 100.00 | 1.00 |
| | 16 | 0.00 | 0.00 | 100.00 | 1.00 | 0.00 | 0.00 | 100.00 | 1.00 | 0.00 | 0.00 | 100.00 | 1.00 | 0.00 | 0.00 | 100.00 | 1.00 |

We found 14 and 82 duplicates in the training sets of CIFAR100 and CelebA, respectively. The testing sets contain 2 and 7 duplicates, respectively. Additionally, 9 of the 14 duplicates in the CIFAR100 training set, as well as 51 of the 82 duplicates in the CelebA training set occur with different labels. Both duplicates in the testing set of CIFAR100 are differently labeled each time.

To our surprise, we also found that the training and testing sets of CIFAR100 and CelebA contain identical images. There are 10 images that occur in both the training and testing set of CIFAR100, 6 of which are differently labeled in the two sets. For CelebA, 11 images are present in both sets, all of which are differently labeled for training and testing.

Table 3 provides a list of all duplicates we found in the CIFAR100 dataset with their IDs and labels.

## 5 Limitations

- We did not discuss the common scenario, where the client does not perform the training on the entire training set, but randomly selects the training samples.
  However, this issue can be mitigated by performing a sufficient number of queries to the client, ensuring membership in at least one random subset with large probability. As our method work with perfect accuracy, this repeated application does not degrade performance.
- We did not conduct investigations into the effect of regularization techniques (e.g. dropout, batch normalization) on the performance of the attack described in this paper.
  However, we remark, that dropout can be easily addressed by using multiple queries. Given enough queries, it can be guaranteed that all necessary neurons are active at the same time with high probability.
- We did not consider possible countermeasures. In particular, Differential Privacy (DP) techniques, where the client adds noise to the parameter vector $\phi$ were not analyzed. These would certainly impact our membership inference strategy. However, the tradeoff between performance and privacy in DP is universal and independent of the specific membership inference attack. The reader is therefore referred to the relevant literature on DP.
- Regarding the model architecture, our method heavily relies on the ReLU activation function and no other activation function was considered. A mild condition is also imposed on the model architecture in that it needs to be possible to embed the structure in Fig. 1 in the model. This requires a depth of at least two layers.
We believe that these requirements are fairly mild considering the ubiquity of the ReLU activation function. Moreover, similar strategies might be possible if other activation functions with piece-wise constant derivatives are used.

- We did not spend any efforts on obfuscation techniques that conceal our attack from the client. Neither did we investigate possible methods of detecting our membership inference attack, for two reasons:

1) The attacker-detector dynamic quickly leads to an "arms-race" where it is certainly easy to provide a detection algorithm to detect the exact structure depicted in 1. However, then an attacker can modify and obfuscate that model structure in some way, which can in turn be detected. This cycle may continue indefinitely.

2) The practical relevance may not be high, as the user is rarely in control of the source code and the exact behavior of the FL client application. In fact, most commonly, the same entity operates the CS and provides the client application binary. In this case, that entity has complete knowledge of any detection code on the client side and will thus win the arms-race alluded to above.

- Other optimizers, apart from SGD and Adam were not studied and are left for future work.
- Similarly, we did not consider FL strategies other than FedAvg, containing FedSGD as a special case.
  However, as long as a optimization relies on gradient computation and the parameter vector is exchanged between the CS and the client, our attacker should be viable.
- We considered only one client. As pointed out in Section 1.1, this is not a limitation for the strategy from the attacker's point of view, because the CS can simply target one client at a time, crafting the parameters sent to that client accordingly. We thus did not study the impact of our method on the legitimate FL system training.
- We did not consider the impact of a Secure Aggregation procedure. However, it has recently been shown that Secure Aggregation does not provide additional privacy when facing an active adversary at the CS [15], [27].

### Acknowledgment

The authors acknowledge TU Wien Bibliothek for financial support through its Open Access Funding Programme.

# References

[1] J. Chen, R. Monga, S. Bengio, and R. Jozefowicz, "Revisiting distributed synchronous sgd," in *International Conference on Learning Representations Workshop Track*, 2016. [Online]. Available: https://arxiv.org/abs/1604.00981 1

[2] B. McMahan, E. Moore, D. Ramage, S. Hampson, and B. A. y Arcas, "Communication-efficient learning of deep networks from decentralized data," in *Proceedings of the 20th International Conference on Artificial Intelligence and Statistics, AISTATS 2017, 20-22 April 2017, Fort Lauderdale, FL, USA*, ser. Proceedings of Machine Learning Research, A. Singh and X. J. Zhu, Eds., vol. 54. PMLR, 2017, pp. 1273–1282. [Online]. Available: http://proceedings.mlr.press/v54/mcmahan17a.html 1

[3] E. Bagdasaryan, A. Veit, Y. Hua, D. Estrin, and V. Shmatikov, "How to backdoor federated learning," in *The 23rd International Conference on Artificial Intelligence and Statistics, AISTATS 2020, 26-28 August 2020, Online [Palermo, Sicily, Italy]*, ser. Proceedings of Machine Learning Research, S. Chiappa and R. Calandra, Eds., vol. 108. PMLR, 2020, pp. 2938–2948. [Online]. Available: http://proceedings.mlr.press/v108/bagdasaryan20a.html 2

[4] L. Melis, C. Song, E. D. Cristofaro, and V. Shmatikov, "Exploiting unintended feature leakage in collaborative learning," in *2019 IEEE Symposium on Security and Privacy, SP 2019, San Francisco, CA, USA, May 19-23, 2019*. IEEE, 2019, pp. 691–706. [Online]. Available: https://doi.org/10.1109/SP.2019.00029 2

[5] J. Geiping, H. Bauermeister, H. Dröge, and M. Moeller, "Inverting gradients - how easy is it to break privacy in federated learning?" in *Advances in Neural Information Processing Systems 33: Annual Conference on Neural Information Processing Systems 2020, NeurIPS 2020, December 6-12, 2020, virtual*, H. Larochelle, M. Ranzato, R. Hadsell, M. Balcan, and H. Lin, Eds., 2020. [Online]. Available: https://proceedings.neurips.cc/paper/2020/hash/c4ede56bbd98819ae6112b20ac6bf145-Abstract.html 2

[6] W. Wei, L. Liu, M. Loper, K.-H. Chow, M. E. Gursoy, S. Truex, and Y. Wu, "A framework for evaluating client privacy leakages in federated learning," in *Computer Security – ESORICS 2020: 25th European Symposium on Research in Computer Security, ESORICS 2020, Guildford, UK, September 14–18, 2020, Proceedings, Part I*. Berlin, Heidelberg: Springer-Verlag, 2020, pp. 545–566. [Online]. Available: https://doi.org/10.1007/978-3-030-58951-6_27 2

[7] H. Hu, Z. Salcic, L. Sun, G. Dobbie, and X. Zhang, "Source inference attacks in federated learning," in *IEEE International Conference on Data Mining, ICDM 2021, Auckland, New Zealand, December 7-10, 2021*, J. Bailey, P. Miettinen, Y. S. Koh, D. Tao, and X. Wu, Eds. IEEE, 2021, pp. 1102–1107. [Online]. Available: https://doi.org/10.1109/ICDM51629.2021.00129 2

[8] L. Zhu, Z. Liu, and S. Han, "Deep leakage from gradients," in *Advances in Neural Information Processing Systems 32: Annual Conference on Neural Information Processing Systems 2019, NeurIPS 2019, December 8-14, 2019, Vancouver, BC, Canada*, H. M. Wallach, H. Larochelle, A. Beygelzimer, F. d'Alché-Buc, E. B. Fox, and R. Garnett, Eds., 2019, pp. 14747–14756. [Online]. Available: https://proceedings.neurips.cc/paper/2019/hash/60a6c4002cc7b29142def8871531281a-Abstract.html 2

[9] B. Zhao, K. R. Mopuri, and H. Bilen, "iDLG: Improved deep leakage from gradients," *CoRR*, vol. abs/2001.02610, 2020. [Online]. Available: http://arxiv.org/abs/2001.02610 2

[10] Z. Wang, M. Song, Z. Zhang, Y. Song, Q. Wang, and H. Qi, "Beyond inferring class representatives: User-level privacy leakage from federated learning," in *2019 IEEE Conference on Computer Communications, INFOCOM 2019, Paris, France, April 29 - May 2, 2019*. IEEE, 2019, pp. 2512–2520. [Online]. Available: https://doi.org/10.1109/INFOCOM.2019.8737416 2

[11] M. Nasr, R. Shokri, and A. Houmansadr, "Comprehensive privacy analysis of deep learning: Passive and active white-box inference attacks against centralized and federated learning," in *2019 IEEE Symposium on Security and Privacy, SP 2019, San Francisco, CA, USA, May 19-23, 2019*. IEEE, 2019, pp. 739–753. [Online]. Available: https://doi.org/10.1109/SP.2019.00065 2

[12] F. Boenisch, A. Dziedzic, R. Schuster, A. S. Shamsabadi, I. Shumailov, and N. Papernot, "When the curious abandon honesty: Federated learning is not private," in *8th IEEE European Symposium on Security and Privacy, EuroS&P 2023, Delft, Netherlands, July 3-7, 2023*. IEEE, 2023, pp. 175–199. [Online]. Available: https://doi.org/10.1109/EuroSP57164.2023.00020 2, 4

[13] L. H. Fowl, J. Geiping, W. Czaja, M. Goldblum, and T. Goldstein, "Robbing the fed: Directly obtaining private data in federated learning with modified models," in *The Tenth International Conference on Learning Representations, ICLR 2022, Virtual Event, April 25-29, 2022*. OpenReview.net, 2022. [Online]. Available: https://openreview.net/forum?id=fwzUgo0FM9v 2, 4

[14] Y. Wen, J. Geiping, L. Fowl, M. Goldblum, and T. Goldstein, "Fishing for user data in large-batch federated learning via gradient magnification," in *International Conference on Machine Learning, ICML 2022, 17-23 July 2022, Baltimore, Maryland, USA*, ser. Proceedings of Machine Learning Research, K. Chaudhuri, S. Jegelka, L. Song, C. Szepesvári, G. Niu, and S. Sabato, Eds., vol. 162. PMLR, 2022, pp. 23 668–23 684. [Online]. Available: https://proceedings.mlr.press/v162/wen22a.html 2

[15] D. Pasquini, D. Francati, and G. Ateniese, "Eluding secure aggregation in federated learning via model inconsistency," in *Proceedings of the 2022 ACM SIGSAC Conference on Computer and Communications Security, CCS 2022, Los Angeles, CA, USA, November 7-11, 2022*, H. Yin, A. Stavrou, C. Cremers, and E. Shi, Eds. ACM, 2022, pp. 2429–2443. [Online]. Available: https://doi.org/10.1145/3548606.3560557 2, 4, 5

[16] Y. Gu, Y. Bai, and S. Xu, "CS-MIA: membership inference attack based on prediction confidence series in federated learning," *J. Inf. Secur. Appl.*, vol. 67, p. 103201, 2022. [Online]. Available: https://doi.org/10.1016/j.jisa.2022.103201 2

[17] A. Salem, Y. Zhang, M. Humbert, P. Berrang, M. Fritz, and M. Backes, "Ml-leaks: Model and data independent membership inference attacks and defenses on machine learning models," in *26th Annual Network and Distributed System Security Symposium, NDSS 2019, San Diego, California, USA, February 24-27, 2019*. The Internet Society, 2019. [Online]. Available: https://www.ndss-symposium.org/wp-content/uploads/2019/02/ndss2019_03A-1_Salem_paper.pdf 2

[18] R. Shokri, M. Stronati, C. Song, and V. Shmatikov, "Membership inference attacks against machine learning models," in *2017 IEEE Symposium on Security and Privacy, SP 2017, San Jose, CA, USA, May 22-26, 2017*. IEEE Computer Society, 2017, pp. 3–18. [Online]. Available: https://doi.org/10.1109/SP.2017.41 2

[19] C. Song, T. Ristenpart, and V. Shmatikov, "Machine learning models that remember too much," in *Proceedings of the 2017 ACM SIGSAC Conference on Computer and Communications Security, CCS 2017, Dallas, TX, USA, October 30 - November 03, 2017*, B. Thuraisingham, D. Evans, T. Malkin, and D. Xu, Eds. ACM, 2017, pp. 587–601. [Online]. Available: https://doi.org/10.1145/3133956.3134077 2

[20] K. A. Bonawitz, V. Ivanov, B. Kreuter, A. Marcedone, H. B. McMahan, S. Patel, D. Ramage, A. Segal, and K. Seth, "Practical secure aggregation for federated learning on user-held data," in *NIPS Workshop on Private Multi-Party Machine Learning*, 2016. [Online]. Available: https://arxiv.org/abs/1611.04482 2

[21] S. Hardy, W. Henecka, H. Ivey-Law, R. Nock, G. Patrini, G. Smith, and B. Thorne, "Private federated learning on vertically partitioned data via entity resolution and additively homomorphic encryption," *CoRR*, vol. abs/1711.10677, 2017. [Online]. Available: http://arxiv.org/abs/1711.10677 2

[22] X. Yang, Y. Feng, W. Fang, J. Shao, X. Tang, S.-T. Xia, and R. Lu, "An accuracy-lossless perturbation method for defending privacy attacks in federated learning," in *Proceedings of the ACM Web Conference 2022*, ser. WWW '22. New York, NY, USA: Association for Computing Machinery, 2022, pp. 732–742. [Online]. Available: https://doi.org/10.1145/3485447.3512233 2

[23] D. P. Kingma and J. Ba, "Adam: A method for stochastic optimization," in *3rd International Conference on Learning Representations, ICLR 2015, San Diego, CA, USA, May 7-9, 2015, Conference Track Proceedings*, Y. Bengio and Y. LeCun, Eds., 2015. [Online]. Available: http://arxiv.org/abs/1412.6980 2

[24] D. J. Beutel, T. Topal, A. Mathur, X. Qiu, T. Parcollet, and N. D. Lane, "Flower: A friendly federated learning research framework," *CoRR*, vol. abs/2007.14390, 2020. [Online]. Available: https://arxiv.org/abs/2007.14390 4

[25] Z. Liu, P. Luo, X. Wang, and X. Tang, "Deep learning face attributes in the wild," in *2015 IEEE International Conference on Computer Vision, ICCV 2015, Santiago, Chile, December 7-13, 2015*. IEEE Computer Society, 2015, pp. 3730–3738. [Online]. Available: https://doi.org/10.1109/ICCV.2015.425 4

[26] B. Barz and J. Denzler, "Do we train on test data? purging CIFAR of near-duplicates," *J. Imaging*, vol. 6, no. 6, p. 41, 2020. [Online]. Available: https://doi.org/10.3390/jimaging6060041 5

[27] M. Lam, G. Wei, D. Brooks, V. J. Reddi, and M. Mitzenmacher, "Gradient disaggregation: Breaking privacy in federated learning by reconstructing the user participant matrix," in *Proceedings of the 38th*

*International Conference on Machine Learning, ICML 2021, 18-24 July 2021, Virtual Event*, ser. Proceedings of Machine Learning Research, M. Meila and T. Zhang, Eds., vol. 139. PMLR, 2021, pp. 5959–5968. [Online]. Available: http://proceedings.mlr.press/v139/lam21b.html 5

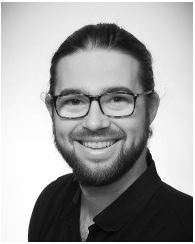

**Georg Pichler** received his M.Sc. and Ph.D. degrees in electrical engineering from the Vienna University of Technology in 2013 and 2017, respectively. He is currently working at the Vienna University of Technology as a post-doctoral research assistant on problems in machine learning, privacy, information theory and related topics.

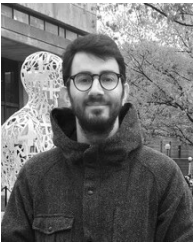

**Marco Romanelli** is a Research Associate at NYU Tandon School of Engineering: Electrical and Computer Engineering. He received his B.Sc. and M.Sc. from the University of Siena, Siena, Italy, and his Ph.D. from École Polytechnique and Inria, Paris, France. His Ph.D. thesis was awarded the UFI Vinci Award. His interests lie at the intersection between information theory and machine learning, with a particular interest in privacy, security, and trustworthy AI. He has authored papers and served as a reviewer for top-tier conferences such as NeurIPS, ACM CCS, and IEEE CSF.

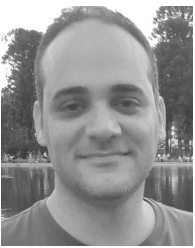

**Leonardo Rey Vega** received the M.Sc. (with honors) and Ph.D. degrees in Electrical Engineering from the University of Buenos Aires (Argentina) in 2004 and 2010, respectively. In 2007 and 2008 he was invited at the INRS-EMT in Montreal, Canada and in the first semester 2012 he was a visitor at the Department of Telecommunications at SUPELEC, France. He is currently an Associate Professor at the University of Buenos Aires. Prof. Rey Vega is a researcher at the National Council for Scientific and Technological Research (CONICET) in Argentina. His research interests include information theory, machine learning and statistical signal processing.

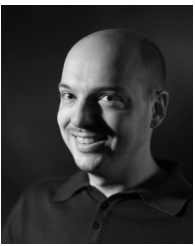

**Pablo Piantanida** (Senior Member, IEEE) received the B.Sc. degree in electrical engineering and the M.Sc. degree from the University of Buenos Aires, Argentina, in 2003, and the Ph.D. degree from Université Paris-Sud, Orsay, France, in 2007. He is currently Professor with CentraleSupélec within Université Paris-Saclay at the International Laboratory on Learning Systems (ILLS) together with CNRS, McGill, ETS and Mila. From 2019 to 2021, he was also an associate member of Comète Inria research team (Lix - Ecole Polytechnique). In 2018 and 2019, he was vising professor at the Montreal Institute for Learning Algorithms (Mila) and Laboratoire de Mathématiques d'Orsay (LMO). His research interests include information theory, machine learning, security of learning systems and the secure processing of information and applications to computer vision, health, natural language processing, among others. He has served as the General Co-Chair for the 2019 IEEE International Symposium on Information Theory (ISIT). He served as an Associate Editor for the IEEE TRANSACTIONS ON INFORMATION FORENSICS AND SECURITY and Editorial Board of Section "Information Theory, Probability and Statistics" for Entropy. He is member of the IEEE Information Theory Society Conference Committee.

TABLE 3
Duplicates in the CIFAR100 dataset.

| Image | Training (ID:Label) | Testing (ID:Label) |
|---|---|---|
| 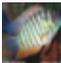 | 4348:`aquarium_fish`, 30931:`aquarium_fish` | – |
| 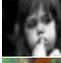 | 8393:`girl`, 36874:`baby` | – |
| 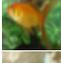 | 8599:`aquarium_fish`, 22657:`aquarium_fish` | – |
| 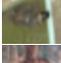 | 9012:`otter`, 31128:`seal` | – |
| 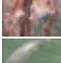 | 16646:`mouse`, 31828:`shrew` | – |
| 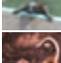 | 17688:`seal`, 41874:`otter` | – |
| 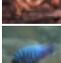 | 18461:`snake`, 46752:`worm` | – |
| 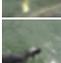 | 20635:`aquarium_fish`, 32666:`aquarium_fish` | – |
| 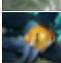 | 23947:`seal`, 33638:`otter` | – |
| 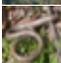 | 25218:`aquarium_fish`, 46851:`aquarium_fish` | – |
| 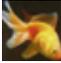 | 27737:`snake`, 47636:`worm` | – |
| 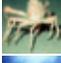 | 28293:`aquarium_fish`, 41860:`aquarium_fish` | – |
| 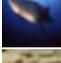 | 30418:`spider`, 47806:`crab` | – |
| 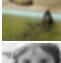 | 31227:`whale`, 34187:`shark` | – |
| 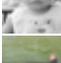 | – | 3438:`otter`, 7715:`seal` |
| 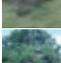 | – | 4654:`girl`, 6709:`baby` |
| 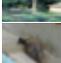 | 24083:`otter` | 1100:`seal` |
| 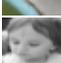 | 32205:`oak_tree` | 1357:`willow_tree` |
| 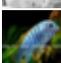 | 28538:`seal` | 2615:`otter` |
| 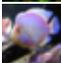 | 30697:`girl` | 2996:`baby` |
| 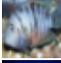 | 28189:`aquarium_fish` | 3499:`aquarium_fish` |
| 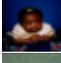 | 21619:`aquarium_fish` | 3932:`aquarium_fish` |
| 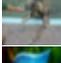 | 29363:`aquarium_fish` | 4716:`aquarium_fish` |
| 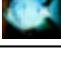 | 25390:`girl` | 5564:`baby` |
| 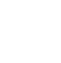 | 37654:`otter` | 6526:`seal` |
| 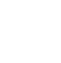 | 15597:`aquarium_fish` | 9524:`aquarium_fish` |